\documentclass[letterpaper, 10 pt, conference]{ieeeconf}  % Comment this line out if you need a4paper

\IEEEoverridecommandlockouts                              % This command is only needed if 
\usepackage{amsmath} % assumes amsmath package installed
\usepackage{amssymb}  % assumes amsmath package installed
\usepackage{todonotes}
\let\labelindent\relax % deactivate ieee labelindent that conflicts with enumitem
\usepackage{enumitem}
\usepackage{mathtools}
\usepackage{rotating}
\usepackage{multirow}
\usepackage{siunitx}
\usepackage{caption}
\usepackage{subcaption}
\usepackage{float}
\usepackage{hyperref}
\usepackage{pifont}
\usepackage{cite}

\usepackage{booktabs} %nice table design

\usepackage{algorithmic} %to display algorithms
\usepackage{algorithm2e}
\graphicspath{{graphics/}}

\usepackage{tabto} % to use tabstops in lists

\usetikzlibrary{matrix, positioning, calc}
\tikzset{
  quadratic/.style={
    to path={
      (\tikztostart) .. controls
      ($#1!1/3!(\tikztostart)$) and ($#1!1/3!(\tikztotarget)$)
      .. (\tikztotarget)
    }
  }
}

\hypersetup{
    colorlinks=true,
    linkcolor=black,
    filecolor=magenta,      
    urlcolor=cyan,
    pdftitle={SnowyLane},
    pdfpagemode=FullScreen,
    }

\title{\LARGE \bf
SnowyLane: Robust Lane Detection on Snow-covered Rural Roads Using Infrastructural Elements 
}

\author{Jörg Gamerdinger$^{1}$, Benedict Wetzel$^{1}$, Patrick Schulz$^{2}$, Sven Teufel$^{1}$, and Oliver Bringmann$^{1}$
\thanks{$^{1}$University of T\"ubingen, Faculty of Science, Department of Computer Science, Embedded Systems Group 
\tt\small {\{joerg.gamerdinger, benedict.wetzel, sven.teufel, oliver.bringmann\} @uni-tuebingen.de}}
\thanks{$^{2}$ Forschungszentrum Informatik (FZI) Karlsruhe 
\tt\small {schulz@fzi.de}}
}%

\begin{document}
\maketitle
\thispagestyle{empty}
\pagestyle{empty}

\begin{abstract}

Lane detection for autonomous driving in snow-covered environments remains a major challenge due to the frequent absence or occlusion of lane markings. In this paper, we present a novel, robust and realtime capable approach that bypasses the reliance on traditional lane markings by detecting roadside features—specifically vertical roadside posts called delineators—as indirect lane indicators. Our method first perceives these posts, then fits a smooth lane trajectory using a parameterized Bézier curve model, leveraging spatial consistency and road geometry. To support training and evaluation in these challenging scenarios, we introduce SnowyLane, a new synthetic dataset containing 80,000 annotated frames capture winter driving conditions, with varying snow coverage, and lighting conditions. Compared to state-of-the-art lane detection systems, our approach demonstrates significantly improved robustness in adverse weather, particularly in cases with heavy snow occlusion. This work establishes a strong foundation for reliable lane detection in winter scenarios and contributes a valuable resource for future research in all-weather autonomous driving. The dataset is available at \url{https://ekut-es.github.io/snowy-lane}

\end{abstract}

%\begin{IEEEkeywords}
%    Lane Detection, Safety Evaluation, Autonomous Driving
%\end{IEEEkeywords}

%%%%%%%%%%%%%%%%%%%%%%%%%%%%%%%%%%%%%%%%%%%%%%%%%%%%%%%%%%%%%%%%%%%%%%%%%%%%%%%%
\section{INTRODUCTION}
\label{sec:intro}
 
\begin{figure}[t]
    \centering
    \includegraphics[width=\linewidth, page=3, trim= 2cm 0cm 2cm 0cm, clip]{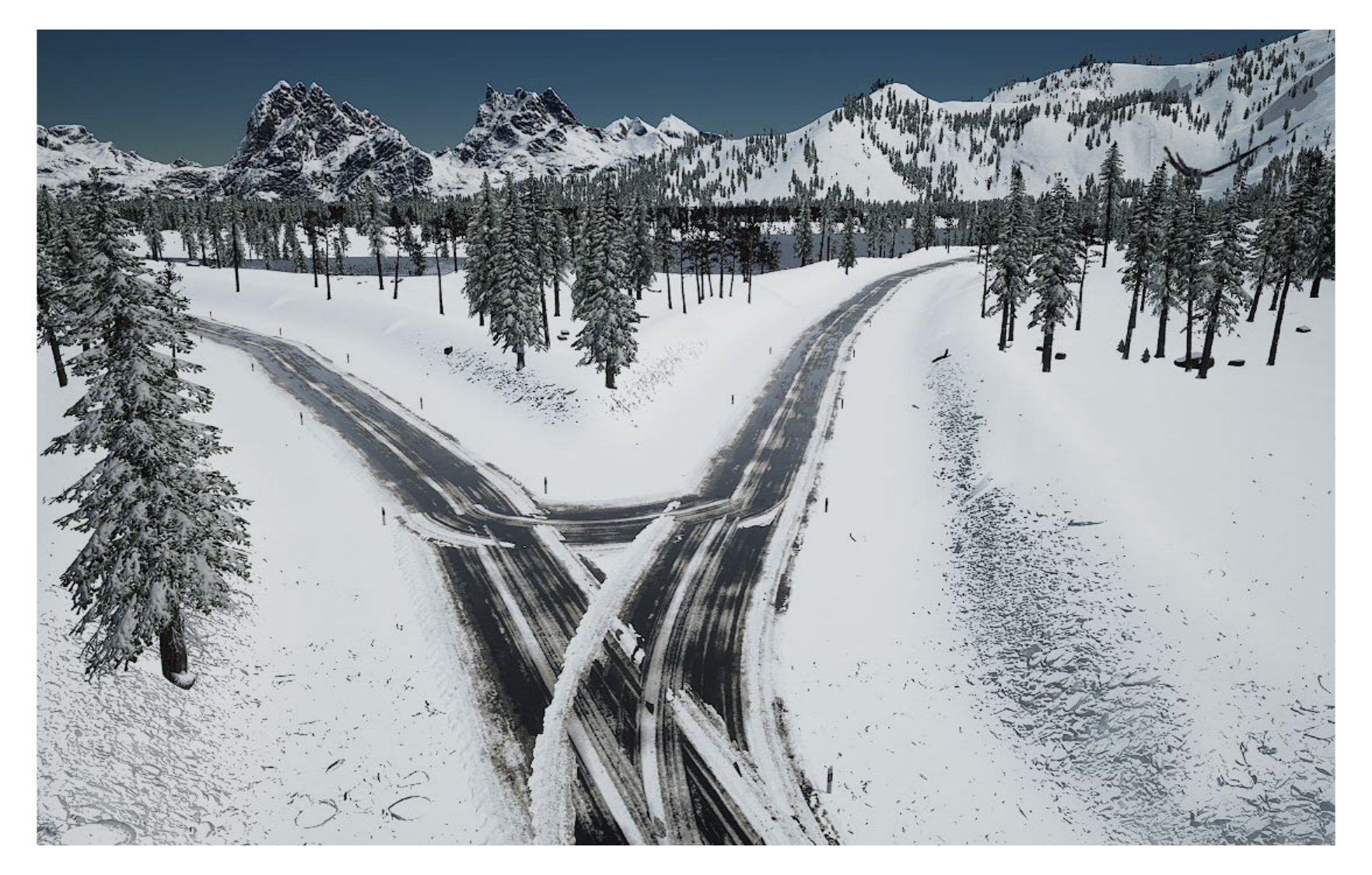}
    \caption{Exemplary scene in a rural environment with partially snow-covered road which can be challenging for lane detection} 
    \label{fig:example}
\end{figure}

The most likely cause of death for young people aged 5 to 29 is road traffic injury~\cite{Death}. Particularly in adverse weather conditions such as rain and snow, the rate of accidents and injuries can increase by \SIrange{70}{80}{\percent}. Automated vehicles are a promising approach to reducing accident rates, as human error is most responsible for fatal road accidents~\cite{EuropeanUnion2019}. 
In order to achieve safe automated driving, a comprehensive perception of the environment is crucial for automated vehicles. However, adverse weather conditions increase the injury rate and also reduce sensing capabilities~\cite{von2019simulating,volk2019towards,Teufel-IV23}. This affects not only object detection but also lane detection, which is necessary for safe trajectory and motion planning. In particular, snow covering the road can severely affect lane detection performance. 
Especially, in rural environments roads show a lower priority for snow removal. However, for rural roads in central Europe, these roads provide vertical posts, called delineators, which are infrastructural elements which can help drivers to perceive the road layout.
Such scenarios occur regularly during winter in northern regions, therefore they must be considered in the development of automated vehicles. 
In order to develop and evaluate robust lane detection in snow-covered environments, suitable datasets are required. Currently available datasets provide adverse weather conditions in the form of rain or fog; however, they lack snow and snow-covered roads. Therefore, we propose SnowyLane, a synthetic dataset with different levels of snow coverage. An exemplary scene is shown in Fig.~\ref{fig:example}. To increase the robustness of lane detectors in snowy environments, we first apply a direct approach, where adverse weather data is included in the lane detector training process. Due to the standardization of the road layout, we propose a novel approach using camera and LiDAR-based object detection methods for infrastructure elements to derive a Bézier curve that can accurately represent the lane markings.

Our main contributions are:
\begin{itemize}
    \item We present SnowyLane, the first large-scale lane detection dataset to include snow-covered roads.
    \item We applied a direct lane detection approach to several state-of-the-art lane detectors in a snow-covered environments, demonstrating their increased robustness and creating a benchmark for SnowyLane.
    \item We propose an efficient and robust lane detection approach for rural roads based on parameterized Bézier curves and infrastructure elements.
\end{itemize}

In Sec.~\ref{sec:related_work} we give an overview on existing datasets and approaches for robust lane detection under adverse weather. The novel SnowyLane dataset is presented in Sec.~\ref{sec:dataset}. Section~\ref{sec:method} describes our approaches for safe and robust lane detection under adverse weather. The results are presented in Sec.~\ref{sec:results}, followed by a conclusion and an outlook to further research.

\begin{figure*}[t]
    \centering
    \includegraphics[width=\linewidth, page=2, trim= 1cm 6cm 1cm 5cm, clip]{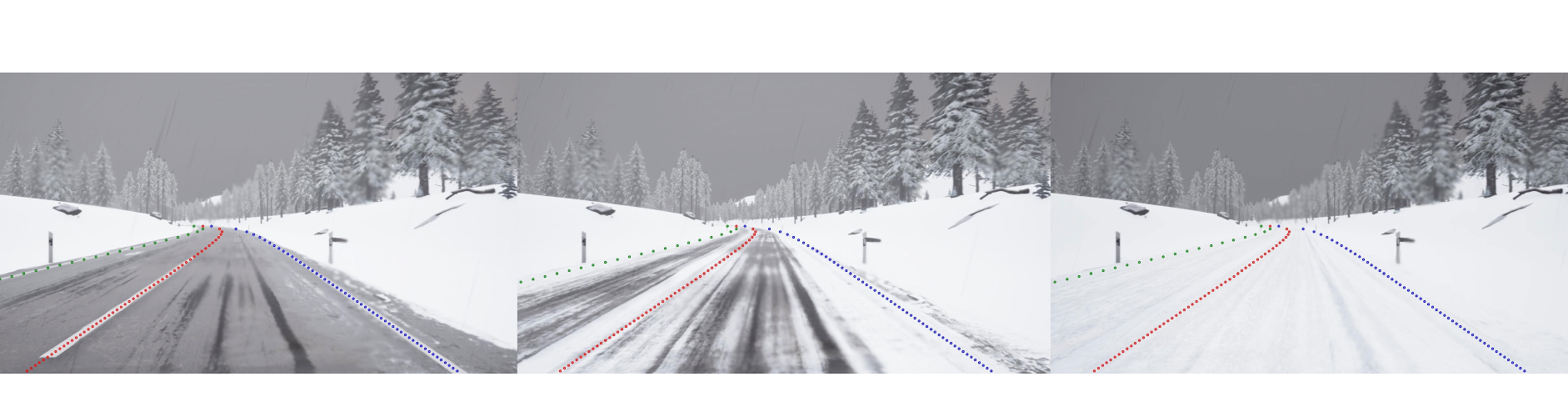}
    \caption{Different snow levels incorporated in the SnowyLane dataset. Low snow (left), medium snow (mid), and high snow (right)} 
    \label{fig:dataset}
\end{figure*}

\section{RELATED WORK}
\label{sec:related_work}

For object detection, robust perception under adverse weather conditions has been widely studied. Volk et al.~\cite{volk2019towards} have shown that incorporating adverse weather images into the training process can significantly improve the performance of camera-based object detection. For LiDAR-based object detection, a similar behavior has been shown by Teufel et al.~\cite{Teufel-IV23}.

When considering lane detection, less research can be observed in the direction of robustness optimization.
Sultana et al.~\cite{sultana2021lane} proposed a method for robust lane detection under adverse weather conditions, where they applied a Hough transform to fit a straight line for lane candidates, which are then filtered using different rules based on previous frames and their angle. Yusuf et al.~\cite{yusuf2020robust} used a convolutional neural network and pre-processing with greyscale conversion and a Gaussian filter to reduce noise. The data from the TuSimple~\cite{TuSimple} dataset contains rain, which is included in the training process, similar to the object detection approach of Volk et al.~\cite{volk2019towards}. A different approach using image optimisation by noise reduction and color space transformation in combination with edge detection and line detection modules with fuzzy logic based adaptive functions was presented by Sang and Norris~\cite{sang2024robust}. All approaches were able to improve perceptual performance over the baseline; however, none of these methods considered snow-covered roads. Vachmanus et al.~\cite{vachmanus2020semantic} used semantic segmentation for road detection under snow coverage. However, their approach only estimates the drivable area and their dataset contained only 400 images, which limits the validity. A second paper was presented by Vachmanus et al.~\cite{vachmanus2022road} for road detection in snowy forest environments. Their approach uses pre-processing and then applies a segmentation of snowy environments and a vanishing point to construct a triangle describing the road. However, their approach relies on the assumption that areas beside the road can be clearly predicted as snow-free due to vegetation. This limits the applicability to real roads, and only the road is segmented, not individual lanes.

In order to develop and test robust lane detection under adverse weather conditions, appropriate datasets are required. Commonly used datasets such as the TuSimple dataset~\cite{TuSimple} or CULane~\cite{pan2018SCNN} lack such weather conditions. Several surveys~\cite{shirke2019lane, zakaria2023lane, he2024monocular} shed light on the topic of lane detection datasets. They show that some datasets such as BDD100k~\cite{yu2020bdd100k} or TTLane~\cite{liang2020lane} include adverse weather in the form of rain and some snow images, but only a few images with falling snow and no snow-covered roads. Only a few images used by Vachmanus et al.~\cite{vachmanus2020semantic, vachmanus2022road} include snow-covered roads; however, they are not publicly available and lack lane information. Another dataset based on ~\cite{sultana2021lane} also includes adverse weather conditions, but the original paper does not provide information on the dataset and it is not publicly available. 

\section{SNOWYLANE DATASET}
\label{sec:dataset}
Due to the lack of appropriate datasets with snow-covered roads, we propose the \textit{SnowyLane} dataset. The synthetic dataset is created using CARLA simulator~\cite{CARLA} and consists of two subsets (with and without delineators). Since the original CARLA maps lack realistic environments, we created a novel map based on German rural roads, including appropriate markings and delineators. 
The map creation was highly automated by using the procedural environment generation by Schulz et al. \cite{schulz2023}. For the OpenDRIVE road network we computed a voronoi graph and transformed its edges into road segments. The landscape has been automatically created with small borders and hills surrounding the road. Forest trees and other foliage are placed procedurally, while still being able to be interchanged for different seasons. To include realistic and adjustable road surfaces, we incorporated the road surface model proposed by Schulz et al. \cite{schulz2024enhancing} and extended it to simulate a snow layer, which combines multiple individually adjustable layers. Additional components, e.g., delineators and accumulated snow that builds up from snowplows, can be automatically placed along the road.

\begin{figure}[t]
    \centering
    \includegraphics[width=\linewidth, trim= 0cm 0cm 0cm 0cm, clip]{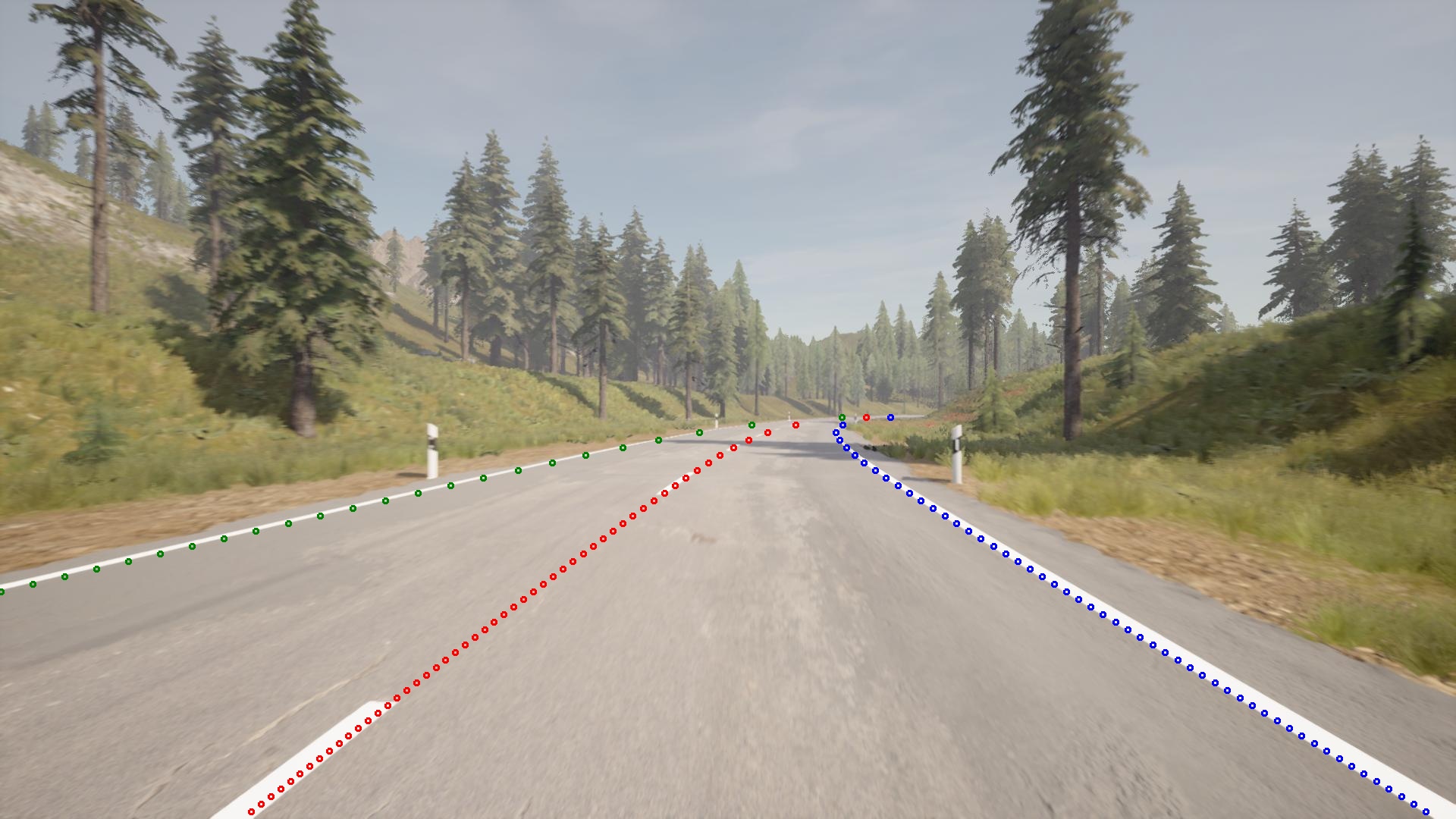}
    \caption{Example image from the summer weather condition SnowyLane subset with delineators}
    \label{fig:summer}
    \vspace*{0mm}
\end{figure}

To investigate the effect of delineators on lane detection in adverse weather conditions, a subset with delineators and a subset without delineators is created for each weather condition.
As a baseline to investigate the differences in perceptual performance between clear summer weather and winter conditions, we created two subsets with clear summer conditions as shown in Fig.~\ref{fig:summer}. For a realistic evaluation, we included three different snow cover intensities for each subset, as shown in Fig.~\ref{fig:dataset}. For the lowest snow cover intensity, the lane markers are still visible. The surroundings of the road are completely covered with snow and the road is wet from melting snow. At medium intensity, the road shows tire tracks from vehicles with less snow cover. The adjacent areas show more coverage with partial coverage of the lane markers at the edge of the road. At high snow intensity, the road is completely covered with snow so that the lane markings and asphalt are no longer visible. To increase the diversity, a randomization of the weather with precipitation, varying cloudiness and different lighting conditions is applied.
The vehicle for recording is equipped with three high-resolution cameras (1920$\times$1080 px.) and a \SI{360}{\degree} 32-layer LiDAR sensor recording with a frequency of \SI{10}{\hertz}. The sensor configuration is shown in Tab.~\ref{tab:sensors}. The semantic segmentation camera is used to fine-tune the ground truth, but extends the usability of the dataset for semantic segmentation of drivable areas. The depth camera is similar to a computed depth image from stereo cameras and allows a 3D position calculation of the delineators in camera-based monocular object detection.

In total, SnowyLane consists of 80,000 frames, of which 20,000 are in summer weather conditions (summer split) and 60,000 with different snow intensities (winter split), 20,000 frames for each of the three snow levels divided into 10,000 frames with delineators and 10,000 frames without delineators. The subsets for different snow intensities are denoted as low snow, medium snow and high snow subset. All snow represents a combination of all snow level intensities.
We provide an 80:10:10 split for training, testing and validation. 
As extensive ground truth, we provide lane marking annotations as 2D polylines with key points as for the highly influential TuSimple dataset~\cite{TuSimple}, which allows for easy usability in different frameworks. In addition, we provide 3D lane information in OpenDrive~\cite{dupuis2010opendrive} and Lanelet~\cite{lanelet} map format, as well as 2D and 3D bounding boxes for the delineators, allowing for camera-based and LiDAR-based detection. Furthermore, we include vehicle state information with velocity, acceleration, orientation, and dimension to enable evaluation with improved metrics for safety assessment such as the LSM~\cite{gamerdinger2024lsm}.

\begin{table}[]
\centering
\caption{Sensor specification}
\begin{tabular}{l  l} \toprule
    Sensor & Specifications \\ \midrule
    1$\times$ RGB Camera  & 1920$\times$1080px, \SI{110}{\degree} FOV \\
    \rule{0pt}{3ex}1$\times$ SemSeg Camera & 1920$\times$1080px, \SI{110}{\degree} FOV \\
    \rule{0pt}{3ex}1$\times$ Depth Camera & 1920$\times$1080px, \SI{110}{\degree} FOV \\
    \rule{0pt}{3ex}1$\times$ LiDAR &\SI{360}{\degree}, \SI{10}{\hertz},\\& \SI{200}{\metre} range, 600k points per \si{\second}\\ \bottomrule
\end{tabular}
\vspace{-1mm}
\label{tab:sensors}
\end{table}

\newpage
\section{ROBUST LANE DETECTION}
\label{sec:method}

\subsection{Direct Lane Detection Approach}
For object detection Volk et al.~\cite{volk2019towards} demonstrated that including adverse weather data in the training process leads to a more robust object detection.
Since adverse weather conditions, especially snow, is rare in different locations, available datasets used for training of neural networks for perception mostly cover only few or no data with these conditions. We used our SnowyLane dataset and the TuRoad lane detection framework~\cite{TuRoadLaneDet} which provides implementations for multiple detectors to investigate the robustness improvement with adverse weather in the training process for lane detection. For all detectors the \SI{80}{\percent} split for training was used. To achieve comparable results, batch size and the number of epochs are equal for each detector. Further augmentation strategies such as noise or transformations are not included to show the robustness increase solely due to adverse weather data.

\subsection{Delineator-based Approach}
\label{subsec:delineator-based}
\begin{figure}[t]
    \centering
    \includegraphics[width=\linewidth, page=1, trim= 3cm 2cm 3cm 2cm, clip]{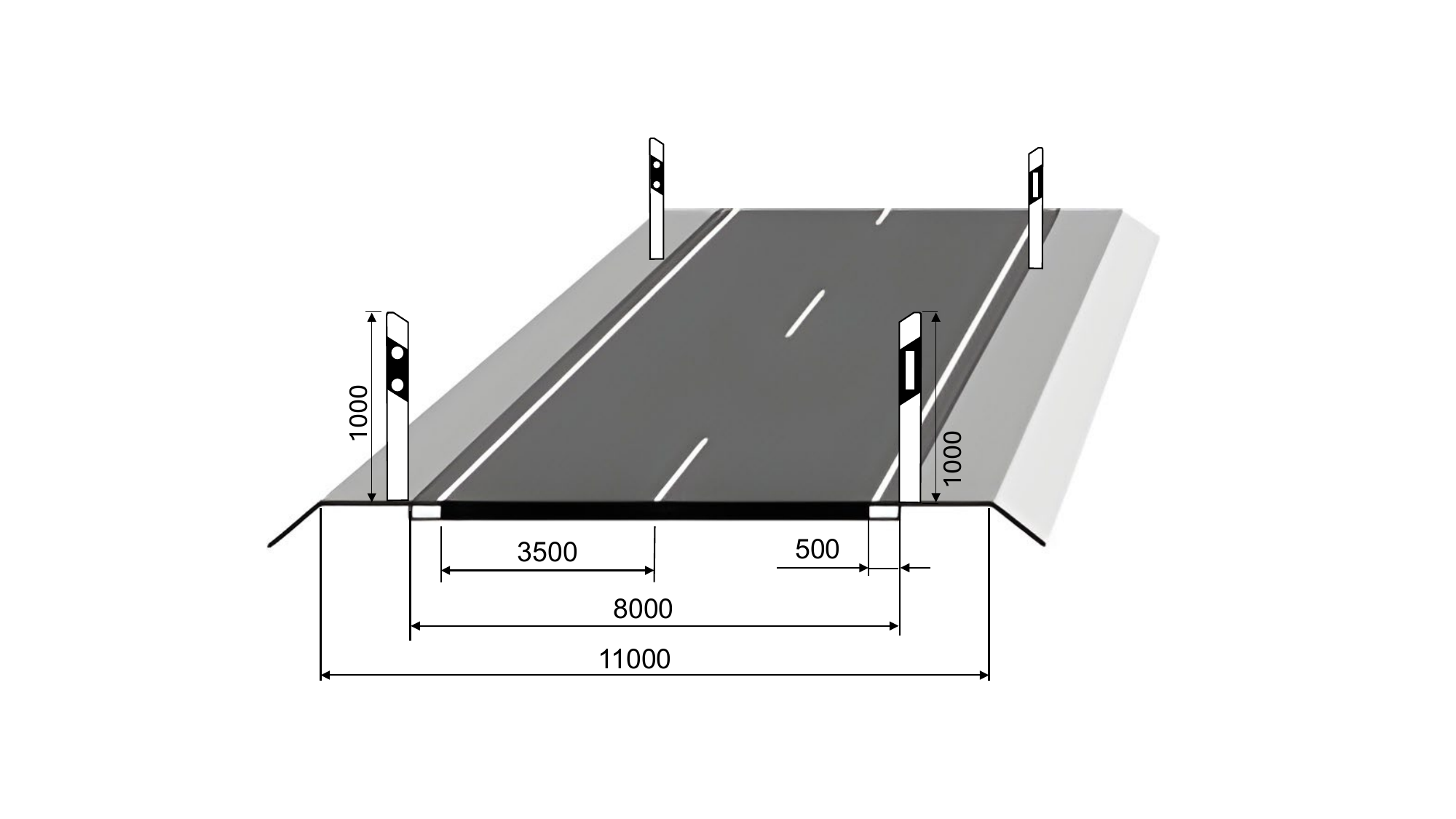}
    \caption{Road layout of German rural roads. Image adapted from~\cite{ral}.   Dimensions in \si{\milli\metre}}
    \label{fig:lane_layout}
\end{figure}

Especially, in rural environments roads show a lower priority for snow removal. However, for German rural roads, these roads provide vertical posts, called delineators, which are infrastructural elements which can help drivers to perceive the road layout. As shown in Fig.~\ref{fig:lane_layout}, these delineators have a standardized height of \SI{1}{\metre} from the bottom of the road, placed $\SI{0.50}{\metre}$ to the outside from the inner lane marker boundary. Additionally, the width of the lane is standardized, which is for this road type $\SI{3.50}{\metre}$~\cite{ral}. Based on the position of the delineators, this allows to estimate the lane markers for each lane with a high precision. Furthermore, the number of lanes can be easily determined by the distance between two corresponding delineators on the left and right side of the road.

For the detection of the delineators, camera or LiDAR-based methods can be applied; here, LiDAR has the advantage that the 3D position is directly known while for the camera a transformation from 2D image coordinates into 3D world coordinates is required which can lead to inaccuracies due to the required depth estimation. 

For road modeling based on some points Bézier curves are well suited~\cite{Aly_2008,dong2024bezierformerunifiedarchitecture2d,feng2023rethinkingefficientlanedetection,WANG2000677}. For a road with $n$ lanes, $n+1$ Bézier curves are required to model each lane marking. The left marking of the road is denoted as $B_l$, the right side $B_r$ and all centerlines are represented as $B_{c,i}$ where $i$ represents the index from left to right.

A 3D Bézier curve can be efficiently described using four points $P_0,P_1,P_2,P_3\in \mathbb R^3$ and is defined as
\begin{equation}
    \mathbf{B(t)} = (1 - t)^3 \mathbf{P_0} + 3(1 - t)^2 t \mathbf{P_1} + 3(1 - t)t^2 \mathbf{P_2} + t^3 \mathbf{P_3}
\end{equation}
with start point $\mathbf{P_0}\in \mathbb R^3$, endpoint $\mathbf{P_3}\in \mathbb{R}^3$ and the two control points $\mathbf{P_1},\mathbf{P_2}\in \mathbb{R}^3$~\cite{foley1996computer}. $t\in [0,1]$ describes the position on the Bézier curve. $\mathbf{P_0}$ and $\mathbf{P_3}$ correspond to the positions of two consecutive delineators.

The two controlpoints $\mathbf{P_1},\mathbf{P_2}$ are derived using the tangents at $\mathbf{P_0}$ and $\mathbf{P_3}$ by the following equations~\cite{foley1996computer}.
\begin{align}
    \mathbf{P_1} &= \mathbf{P_0} + \alpha \cdot \vec{\mathbf{T_0}}\\
    \mathbf{P_2} &= \mathbf{P_3} - \beta \cdot \vec{\mathbf{T_1}}
\end{align}

$\alpha$ and $\beta$ describe a distance which normally corresponds to $\approx \frac{1}{3}$ of the total distance of the Bézier curve. $\vec{T}_0, \vec{T}_1$ represent the direction vectors at $\mathbf{P_0}$ and $\mathbf{P_1}$, respectively.

If more than two consecutive delineators can be perceived, multiple Bézier curves are calculated, each between to consecutive delineators. In this case a second Bézier curve would be described by $P_3, P_4, P_5,P_6\in \mathbb R^3$ with $P_6$ as third delineator position. In order to achieve smooth transitions so called positional continuity and velocity continuity must be fulfilled. Based on Farin~\cite{curves_and_surfaces}, the positional continuity is fulfilled as the endpoint $P_3$ of the first Bézier curve $B_1$ is the startpoint of the second Bézier curve $B_2$. For the velocity continuity the following equation must hold~\cite{curves_and_surfaces}:
\begin{equation}
    \mathbf{P_3} - \mathbf{P_2} = \mathbf{P_4} - \mathbf{P_3}
\end{equation}
This means that the two controlpoints which are next to $P_3$ must be mirrored at $P_3$.
Due to these properties, consecutive Bézier curves allow a smooth modeling of the road layout. After calculating the two Bézier curves for the left $B_l$ and right $B_r$ side of the road, the lane markings must be estimated. For the outer lane markings $B_l$ must be shifted \SI{0.50}{\metre} to the right and $B_r$ \SI{0.50}{\metre} to the left to compensate the offset of the delineator to the lane marking.

A shifting can be performed using Eq.~\eqref{eq:tangent}-~\eqref{eq:offset}.
First, the tangent $T_i$ for all $P_i\in\{P_0 - P_3\}$ describing the Bézier curve must be determined using 
\begin{equation}
    \label{eq:tangent}
    \vec{\mathbf{T}(t)} = \frac{d\mathbf{B}(t)}{dt}.
\end{equation}
Then the normal vector is required, which is calculated by
\begin{equation}
    \label{eq:normal}
    \vec{\mathbf{N}(t)} = \frac{\mathbf{U} \times \vec{\mathbf{T}(t)}}{\left\|\mathbf{U} \times \vec{\mathbf{T}(t)}\right\|}.
\end{equation}
In this case, the left-pointing normal vector is used to shift $B_r$ to the left. For this the global up vector $U$ is used.
\begin{equation}
\mathbf{U} = 
\begin{bmatrix}
0 \\
0 \\
1
\end{bmatrix}
\end{equation}

Finally, the Bézier curve is shifted using
\begin{equation}
    \label{eq:offset}
    \mathbf{B}_{\text{shifted\_left}}(t) = \mathbf{B}(t) + d \cdot \vec{\mathbf{N}(t)}.
\end{equation}
where $d$ represents the shifting distance.
A shifting to the right would be performed by inverting the direction of the normal vector as shown in the following equation.
\begin{equation}
    \label{eq:offset2}
    \mathbf{B}_{\text{shifted\_right}}(t) = \mathbf{B}(t) - d \cdot  \vec{\mathbf{N}(t)}
\end{equation}

Mostly, rural roads consist of two lanes. In this case, the distance between two delineators on the left and right side of the roads is about \SI{8}{\metre} and $B_l$ or $B_r$ can be shifted towards the center of the road by the lane width (here \SI{3.5}{\metre}). A second approach to overcome possible inaccuracies of a single boundary, is to calculate the center points of two corresponding delineators and use these points to determine the third Bézier curve $B_c(t)$ representing the centerline. 
In some cases a third overtaking lane may be present, in this case the distance between the delineator increases which allows to estimate the number of lanes by
\begin{equation}
\label{eq:num_lanes}
    \#lanes = \frac{d_{delineators} - 1.0}{w_l}
\end{equation}
where $d_{delineators}$ describe the distance between the two corresponding delineators on the left and right side of the road and $w_l$ the lane width which depends on the road type. \SI{1.0}{\metre} must be subtracted to exclude the difference between the delineators and the lane markings.
For such cases the Bézier curve of one lane boundary can be shifted by $w_l$ according to the number of lanes using Eq.~\eqref{eq:num_lanes}. This allows for an increased robustness in cases such as multi-lane rural roads or merging lanes.

In some cases the lane prediction distance is limited as the detection is performed using the interpolation of the Bézier curve. Since roads less likely show unpredictable layouts, the modelled 3D curve allows for some extrapolation of the determined Bézier curve in order to increase the detection range. This process must be performed conservatively, as with an increase prediction horizon the deviation in the estimation will become less accurate. Our experiments with all detectors presented in Sec.~\ref{sec:results} on the whole dataset showed that an extrapolation of \SI{10}{\metre} can be used to increase the detection range. Extrapolating higher distances leads to significant errors reducing the results of all incorporated metrics.

\begin{table*}
      \caption{Results for the direct approach using camera-based lane detectors and our delineator-based approach using camera-based (Faster R-CNN, YOLOv5, SMOKE) and LiDAR-based (Voxel R-CNN, PV-RCNN) object detectors.\\ Results are (2D Accuracy, 3D Accuracy, LSM)}
      \centering
      \centering
\renewcommand{\arraystretch}{1.3}
\resizebox{\textwidth}{!}{%
    \begin{tabular}{cccccccccc} \toprule
    & &  \multicolumn{3}{c}{\textbf{Direct Lane Detection Approach}} &  \multicolumn{5}{c}{\textbf{Delineator-based Approach (ours)}}\\
    Dataset &Delineator&SCNN& RESA & UFLD & Faster R-CNN & YOLOv5 & SMOKE& Voxel R-CNN & PV-RCNN \\ \midrule
    Summer only& Yes& (0.92, 0.94, 0.98) & (0.92, 0.93, 0.98)  & (0.82, 0.82, 1.00) & (-, 0.58, 0.97) & (-, 0.58, 0.90) & (-, 0.39, 0.92)& (-, 0.53, 0.64)& (-, 0.15, 0.82)   \\
    \midrule\midrule
    Baseline Low Snow& Yes& (0.56, 0.47, 0.04) & (0.55, 0.62, 0.01)  & (0.63, 0.26, 0.97) & (-, 0.00, 0.00) & (-, 0.00, 0.00)& (-, 0.02, 0.24) & (-, 0.57, 0.63) & (-, 0.61, 0.90)  \\
    Baseline High Snow& Yes& (0.04, 0.00, 0.00) & (0.04, 0.00, 0.00)  & (0.62, 0.14, 0.94) & (-, 0.00, 0.00)& (-, 0.00, 0.00)& (-, 0.03, 0.35) & (-, 0.68, 0.62) & (-, 0.74, 0.90)  \\
    Baseline All Snow& Yes& (0.04, 0.03, 0.00) & (0.06, 0.09, 0.00) & (0.62, 0.14, 0.92) & (-, 0.00, 0.01) &(-, 0.00, 0.00) & (-, 0.02, 0.68) & (-, 0.51, 0.63) & (- 0.56, 0.91)  \\
    \midrule\midrule
    Low Snow& No&(0.92, 0.04, 0.99) & (0.92, 0.05, 0.98)  & (0.82, 0.15, 1.00) & - & - & - & - & - \\
    Medium Snow& No&(0.90, 0.08, 1.00) & (0.90, 0.07, 1.00)  & (0.81, 0.20, 1.00) & - & - & - & - & -  \\
    High Snow& No&(0.85, 0.17, 0.99) & (0.84, 0.18, 0.99)  & (0.68, 0.32, 0.99) & - & - & - & - & - \\
    All Snow& No&(0.87, 0.12, 0.99) & (0.87, 0.13, 0.99)  & (0.75, 0.32, 0.99) & - & - & - & - & - \\
    \midrule\midrule
    Low Snow& Yes&(0.92, 0.06, 0.99) & (0.92, 0.06, 0.99)  & (0.88, 0.18, 1.00) & (-, 0.57, 0.98)& (-, 0.60, 0.91) & (-, 0.19, 0.94)& (-, 0.57, 0.61) & (-, 0.62, 0.91)  \\
    Medium Snow& Yes&(0.90, 0.09, 1.00) & (0.90, 0.08, 0.99)  & (0.79, 0.26, 0.99) & (-, 0.59, 0.97)& (-, 0.59, 0.94)& (-, 0.04, 0.84) & (-, 0.52, 0.61) & (-, 0.52, 0.90)  \\
    High Snow& Yes&(0.75, 0.28, 0.98) & (0.74, 0.29, 0.98)  & (0.64, 0.14, 0.66) & (-, 0.73, 0.95)& (-, 0.74, 0.96)& (-, 0.45, 0.93) &(-, 0.67, 0.59) & (-, 0.74, 0.91) \\
    All Snow& Yes&(0.87, 0.13, 0.98) & (0.87, 0.13, 0.99)  & (0.73, 0.27, 0.99) & (-, 0.57, 0.99)& (-, 0.57, 0.96)& (-, 0.05, 0.81) & (-, 0.50, 0.62) & (-, 0.56, 0.90)   \\
    
    \bottomrule
\end{tabular}
}
\label{tab:results}
\vspace*{4mm}
\end{table*}
\section{EXPERIMENTS AND RESULTS}
\label{sec:results}

\begin{figure*}[t]
\centering
    \begin{subfigure}{0.30\linewidth}
        \includegraphics[width=\linewidth]{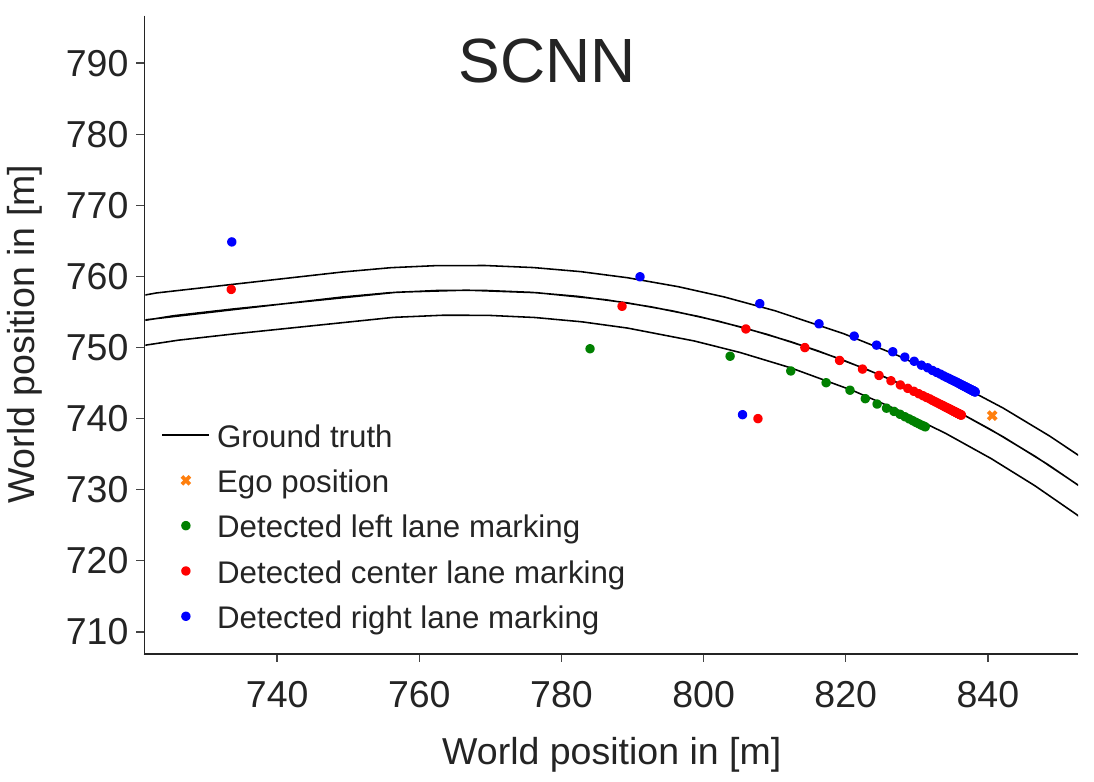}
    \caption{SCNN}
    \end{subfigure}
\hfill
\begin{subfigure}{0.30\linewidth}
        \includegraphics[width=\linewidth]{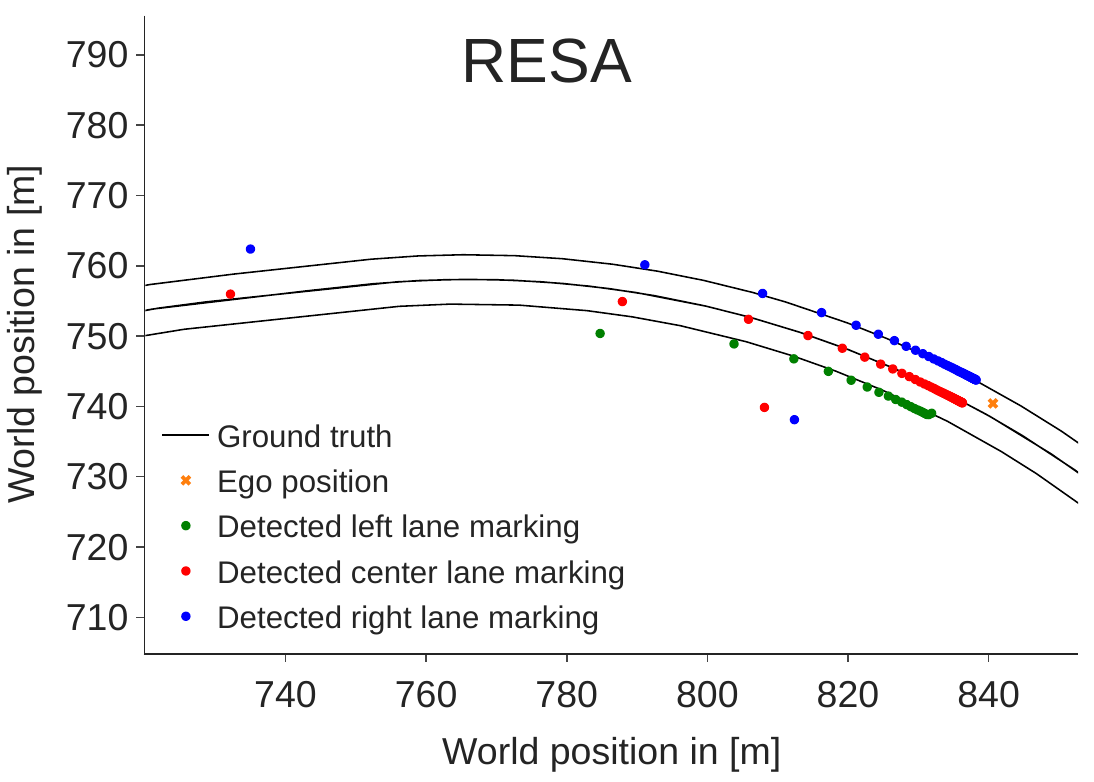}
    \caption{RESA}
    \end{subfigure}
\hfill
\begin{subfigure}{0.30\linewidth}
        \includegraphics[width=\linewidth]{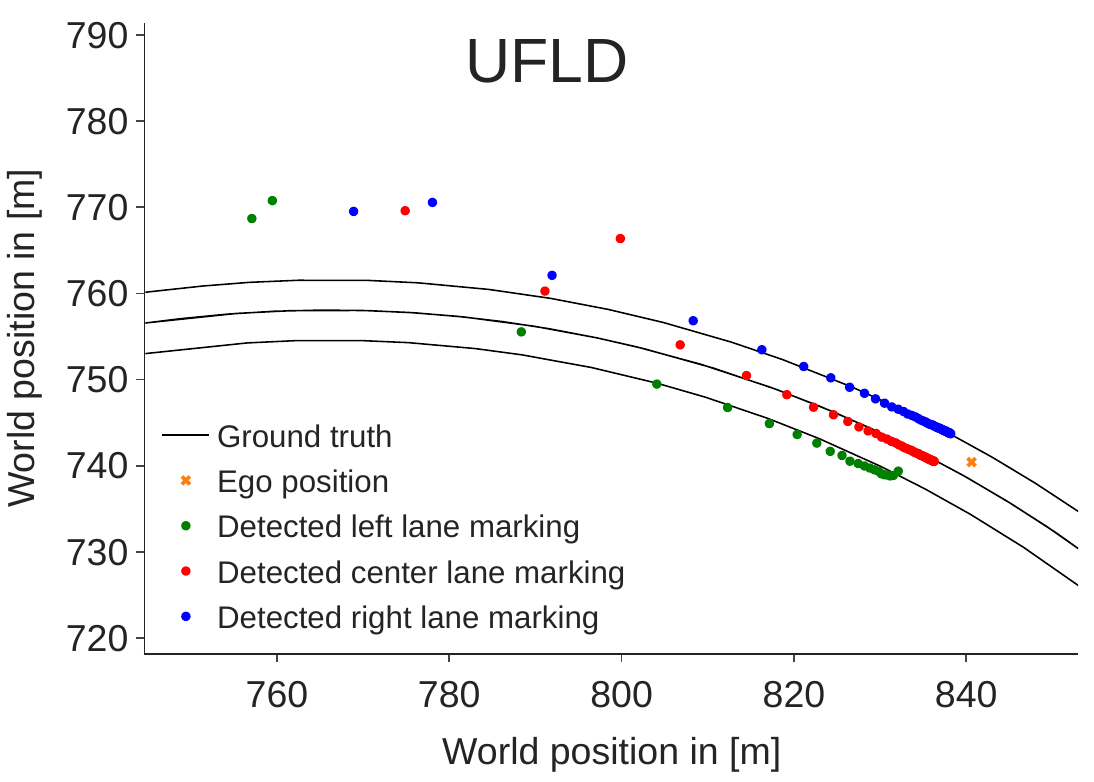}
    \caption{UFLD}
    \end{subfigure}
\hfill
    \begin{subfigure}{0.30\linewidth}
        \includegraphics[width=\linewidth]{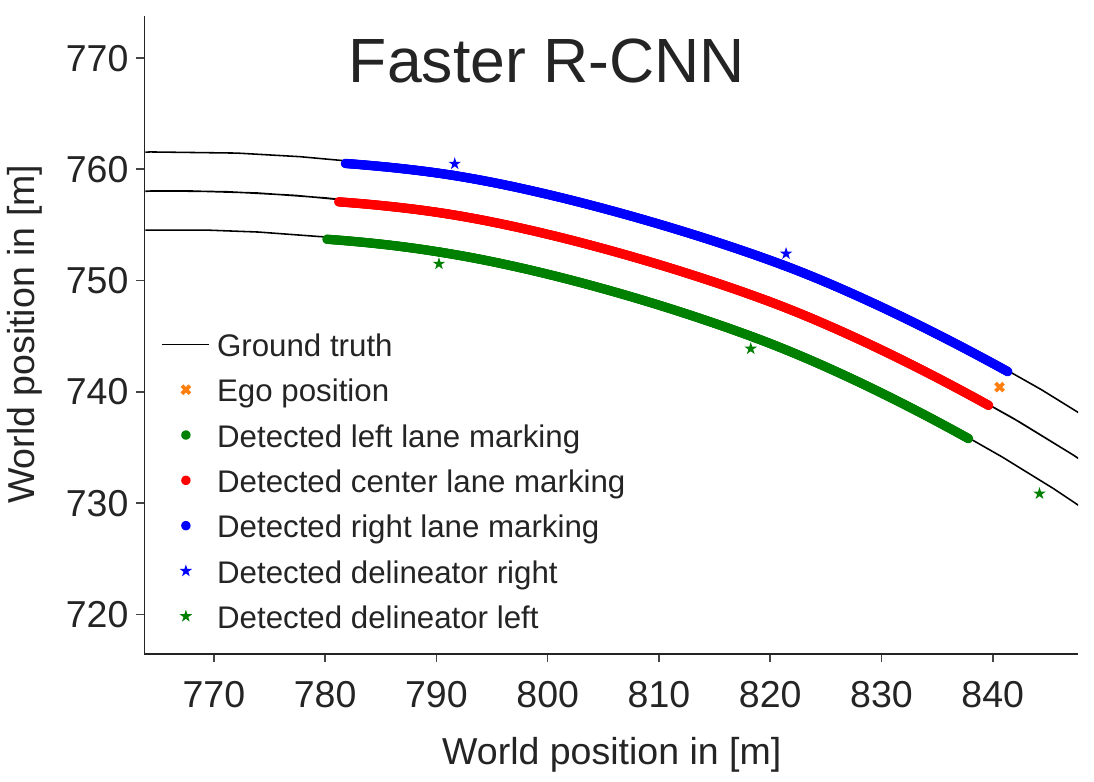}
    \caption{Faster R-CNN}
    \end{subfigure}
\hfill
\begin{subfigure}{0.30\linewidth}
        \includegraphics[width=\linewidth]{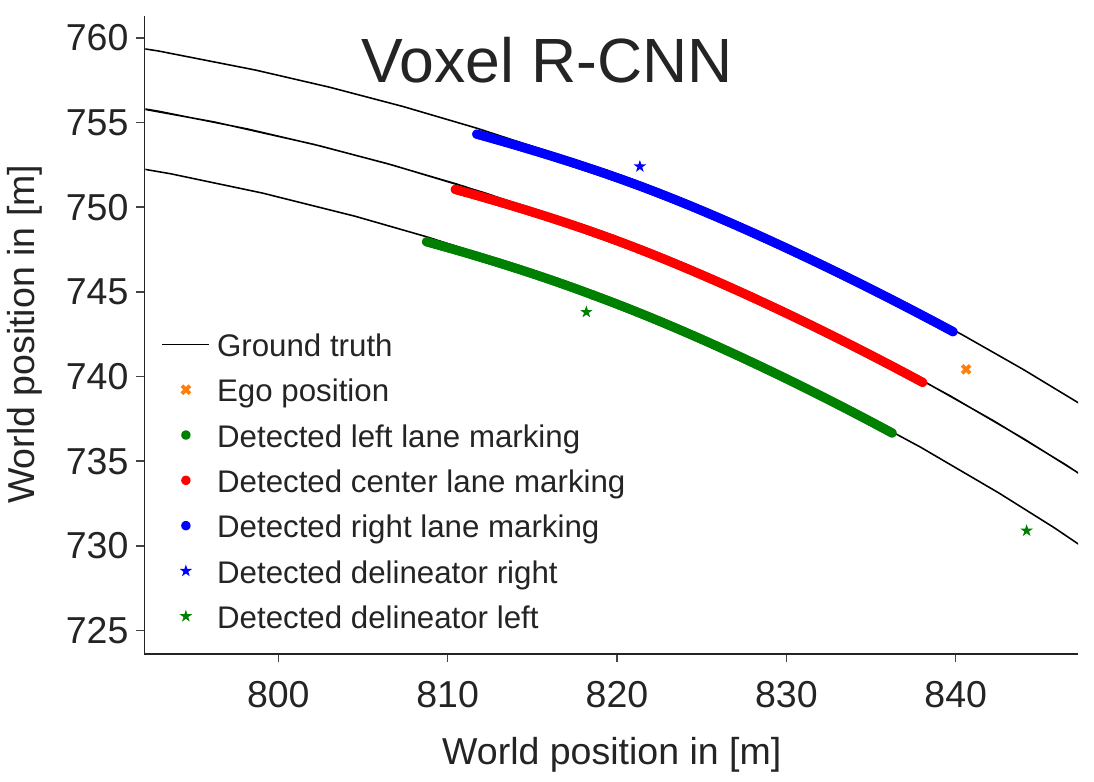}
    \caption{Voxel R-CNN}
    \end{subfigure}
\hfill
    \begin{subfigure}{0.30\linewidth}
        \includegraphics[width=\linewidth]{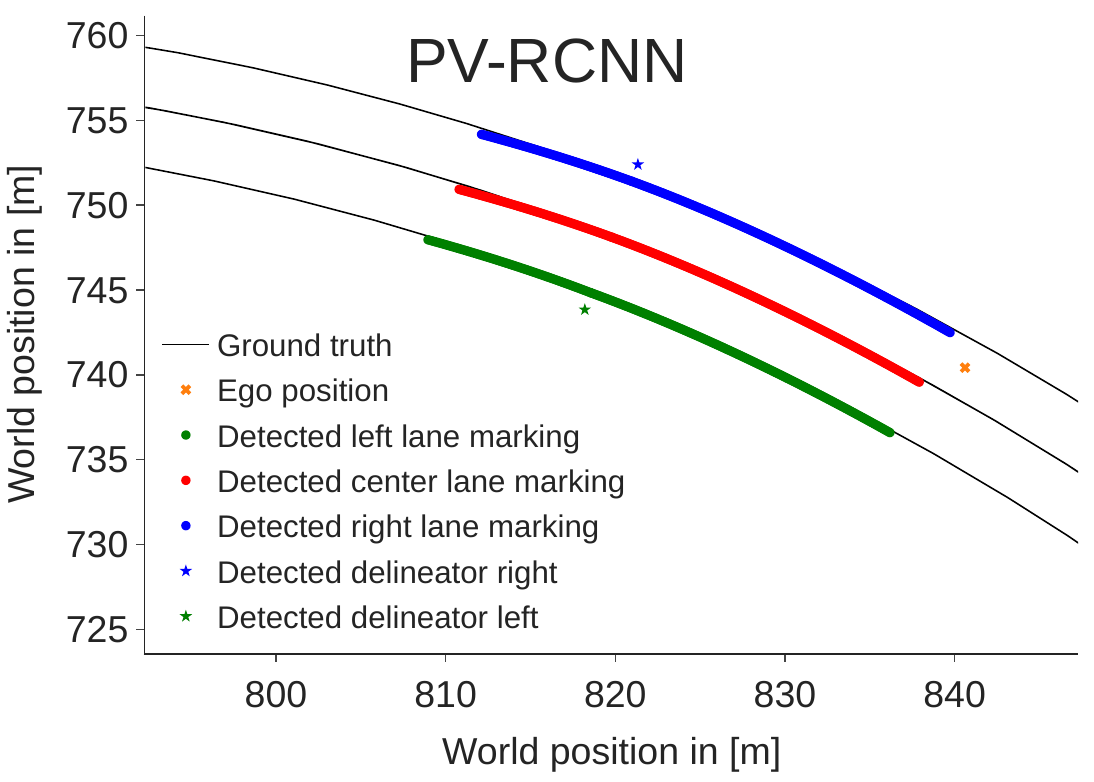}
    \caption{PV-RCNN}
    \end{subfigure}
   
\caption{Exemplary predictions of direct approach with (a) SCNN, (b) RESA and (c) UFLD as well as the delineator-based approach with (d) Faster R-CNN, (e) Voxel R-CNN and (f) PV-RCNN.}
%\vspace*{-5mm}
    \label{fig:examples}
    \end{figure*}

To evaluate the performance and safety of the two approaches, we use the 2D accuracy as for the TuSimple benchmark with a threshold of 5 px. for a correct detection as well as 3D accuracy and the lane safety metric (LSM)~\cite{gamerdinger2024lsm} with a true positive threshold of \SI{0.10}{\metre}. Especially, the LSM allows for a more meaningful statement about the safety which can be achieved in adverse weather using the different methods. An overview of the results is presented in Tab.~\ref{tab:results}. Example predictions are shown in Fig.~\ref{fig:examples}. All detectors are trained on the winter split of the SnowyLane dataset with delineators and evaluated using the three snow coverage intensity subsets introduced in Sec.~\ref{sec:dataset}. Additionally, we evaluate on an all snow subset in which all three snow coverage intensities are incorporated. Furthermore, snowfall augmentation is included using the albumentations framework~\cite{info11020125} for the camera data and the LidarAug framework by Teufel et al.~\cite{lidar-weather} for the LiDAR data. For both methods, we included snowfall rates of about \SIrange{5}{20}{\milli\meter\per\hour} with varying snowflake sizes.

To evaluate the direct approach and create a baseline for the delineator-based approach we used three state-of-the-art lane detectors from the TuRoad framework~\cite{TuRoadLaneDet}. These are SCNN~\cite{pan2018SCNN}, UFLD~\cite{qin2020ultra} and RESA~\cite{zheng2021resa}. 
For the delineator-based approach, object detectors to perceive the delineators are required. For our experiments we used the 2D camera-based object detectors Faster R-CNN~\cite{ren2015faster}, YOLOv5~\cite{yolov5} and the 3D monocular detector SMOKE~\cite{liu2020smoke}. Furthermore, we use LiDAR-based object detectors PV-RCNN~\cite{shi2020pv} and Voxel R-CNN~\cite{deng2021voxel} included in the OpenPCDet framework~\cite{openpcdet2020}.

In order to demonstrate the realtime capability, we evaluate the runtime of the required calculations using the \textit{All Snow} subset with 30k frames and all three snow levels.
These experiments are performed on an AMD Ryzen 9 5950X with \SI{128}{\giga\byte} RAM and a Nvidia RTX4090 GPU. 
\newpage
\subsection{Baseline}
As a baseline for all methods, we trained all detectors using the 10,000 frame summer split of SnowyLane with delineators and evaluate on the low snow, high snow and all snow datasets. Moreover, we created a baseline for the summer dataset with delineators.
For this summer subset, the lane detectors achieve very high results with 2D accuracies between 0.82 for UFLD and 0.92 for the others. Also the 3D accuracy ranges between 0.82 and 0.92. This good detections lead to very high nearly perfect LSM results with 0.98 for SCNN and RESA and 1.00 for UFLD. The delineator-based approach for camera and LiDAR-based detection also performed well with slightly lower values ranging between 0.64 for Voxel R-CNN and 0.97 for Faster R-CNN.

For the baseline, we can observe a drop in performance. While for low snow all lane detectors still achieve a high 2D accuracy and LSM score, the 3D performance drops by about \SI{90}{\percent}. The high 2D accuracy can be explained as the lane markings in the low snow subset are still visible so a few points are detected correctly which leads to a high accuracy. For high and all snow, the performance of SCNN and RESA drops significantly to a 2D accuracy of about 0.05 and 3D accuracy of 0.00-0.09 which also leads to a LSM score of 0.00. Only UFLD is able to still detect the lane properly, reaching 0.62 in 2D accuracy and 0.92 for LSM. Even though there is a lot of variation in snow levels, environmental appearance and road course, the rural roads always consist of 2 lanes with a fixed width which allows the detector to learn this behavior already on the summer images and can then predict this under snowy conditions.

\begin{table*}
      \caption{Results with snowfall augmentations for the direct approach using camera-based lane detectors and our delineator-based approach using camera-based (Faster R-CNN, YOLOv5, SMOKE) and LiDAR-based (Voxel R-CNN, PV-RCNN) object detectors.Results are (2D Accuracy, 3D Accuracy, LSM)}
      \centering
      \centering
\renewcommand{\arraystretch}{1.3}
\resizebox{\textwidth}{!}{%
    \begin{tabular}{cccccccccc} \toprule
    & &  \multicolumn{3}{c}{\textbf{Direct Lane Detection Approach}} &  \multicolumn{5}{c}{\textbf{Delineator-based Approach (ours)}}\\
    Dataset &Delineator&SCNN& RESA & UFLD & Faster R-CNN & YOLOv5& SMOKE & Voxel R-CNN & PV-RCNN \\ \midrule
    Baseline Low Snow& Yes& (0.47, 0.49, 0.03) & (0.53, 0.64, 0.01)  & (0.63, 0.22, 0.96) & (-, 0.00, 0.00) & (-, 0.00, 0.00)& (-, 0.02, 0.56) & (-, 0.58, 0.63) & (-, 0.61, 0.95)   \\
    Baseline High Snow& Yes& (0.03, 0.00, 0.00) & (0.04, 0.00, 0.00)  & (0.62, 0.07, 0.94) & (-, 0.00, 0.00)& (-, 0.00, 0.00)& (-, 0.03, 0.35) & (-, 0.69, 0.63) & (-, 0.70, 0.95)   \\
    Baseline All Snow& Yes& (0.19, 0.49, 0.01) & (0.22, 0.63, 0.00) & (0.62, 0.12, 0.95) & (-, 0.00, 0.00) &(-, 0.00, 0.00) & (-, 0.02, 0.70) & (-, 0.55, 0.64) & (-, 0.56, 0.95)  \\
    \midrule\midrule
    Low Snow& Yes&(0.91, 0.05, 0.99) & (0.92, 0.05, 0.99)  & (0.80, 0.17, 1.00) & (-, 0.57, 0.97)& (-, 0.60, 0.91)& (-, 0.40, 0.92)  & (-, 0.56, 0.65) & (-, 0.61, 0.90)  \\
    Medium Snow& Yes&(0.84, 0.12, 0.99) & (0.84, 0.12, 0.99)  & (0.72, 0.10, 0.99) & (-, 0.09, 0.91)& (-, 0.12, 0.92)& (-, 0.04, 0.75) & (-, 0.51, 0.62) & (-, 0.57, 0.91)   \\
    High Snow& Yes&(0.75, 0.11, 0.99) & (0.73, 0.11, 0.99)  & (0.65, 0.09, 0.98) & (-, 0.09, 0.90)& (-, 0.12, 0.92)& (-, 0.45, 0.92) &(-, 0.71, 0.66) & (-, 0.72, 0.91)\\
    All Snow& Yes&(0.87, 0.10, 0.99) & (0.87, 0.10, 0.99)  & (0.77, 0.14, 0.99) & (-, 0.27, 0.95)& (-, 0.28, 0.91)& (-, 0.04, 0.76) & (-, 0.53, 0.61) & (-, 0.58, 0.91)   \\
    
    \bottomrule
\end{tabular}
}
\label{tab:results_snow_aug}
\vspace*{-4mm}
\end{table*}   

Considering the delineator-based approach, it can be observed that the 2D camera-based methods (Faster R-CNN and YOLO) are not capable of detecting any delineators leading to accuracy and LSM scores of 0.00. In the summer data for training, there is a good contrast between the green environment and white delineators while for testing in winter conditions the white delineators can not be distinguished from the snow-covered environment; hence, a drop in perception performance and safety can be observed. In contrast, the SMOKE 3D object detection achieved LSM scores ranging between 0.24 and 0.68. Under these conditions the delineator-based approach with LiDAR-based object detection performs really well. Especially, using PV-RCNN a 3D accuracy of up to 0.74 with a LSM score of about 0.90 can be achieved, which corresponds to a highly safe perception. 
This experiment clearly demonstrate that lane detection models trained with data in clear summer conditions are not capable to achieve a sufficient prediction in snow-covered environments. Hence, there exists the necessity to improve robustness in lane detection in order to achieve safe autonomous driving under all environmental conditions.

\subsection{Direct Lane Detection Approach}
\label{subsec:train-based}
For the direct approach without poles a high 2D accuracy of about 0.90 for all detectors over all snow intensities can be observed. Considering the 3D accuracy the results are rather low and range between 0.04 for SCNN and 0.32 for UFLD. For 2D accuracy, SCNN and RESA performed best, regarding the 3D accuracy UFLD performed best. In terms of safety, the LSM score is for all detectors and snow levels at about 1.0 which corresponds to a complete safe situation. This can be explained as the lateral deviation is not high enough to decrease the safety score and the longitudinal detection range is sufficient enough as the vehicles speed is only \SI{50}{\kilo\metre\per\hour} which leads to a low required detection range. In order to evaluate if the detectors can learn the lane layout implicitly with poles, we also evaluated this case. Here, similar to without delineators all models achieved similar results. For low snow, medium snow and all snow the 2D accuracy again is about 0.90 while the 3D accuracy ranges between 0.06 and 0.29. Only for the high snow case over all detectors lower results can be observed with a 2D accuracy ranging between 0.64 for UFLD and 0.75 for SCNN. The LSM is also about 1.0 except for UFLD with high snow with 0.66 which corresponds to a moderate risk. In general it can be observed, that the performance decreases for higher snow cover intensities and that the effect of delineators to the lane detector based detection is rather low.
\newpage
\subsection{Delineator-based Approach}
The delineator-based approach is evaluated using 3D accuracy and the LSM. Compared to the direct approach, for the 2D camera-based algorithms an increase of 3D accuracy by up to 0.46 can be achieved for YOLOv5 against RESA as the best direct lane detector in the high snow subset. For the other subsets 3D accuracy is at least 0.57, which is also significantly higher than for the lane detectors with a maximum of 0.29. In terms of safety a slight decrease of less than 0.1 in the LSM score can be observed. However, the LSM score still ranges between 0.91 and 0.99 which corresponds to a completely safe situation. Considering the monocular 3D object detection model SMOKE a 3D accuracy up to 0.45 which is an increase of 0.16 compared to the best lane detection model was achieved. SMOKE also achieved high LSM scores of about 0.80 to 0.95 which can be considered as really safe and comparable to the LiDAR-based methods.

Using LiDAR-based detectors, similar results with 3D accuracies ranging between 0.50 and 0.67 for Voxel-RCNN and 0.52 and 0.74 for PV-RCNN. While Voxel-RCNN due to a lower detection range only achieves LSM scores around 0.60, with PV-RCNN LSM scores of about 0.90 which corresponds to a completely safe situation.

\subsection{Snowfall Augmentation}
\begin{figure}[t]
    \centering
    \includegraphics[width=\linewidth, page=6, trim= 8cm 0cm 8cm 0cm, clip]{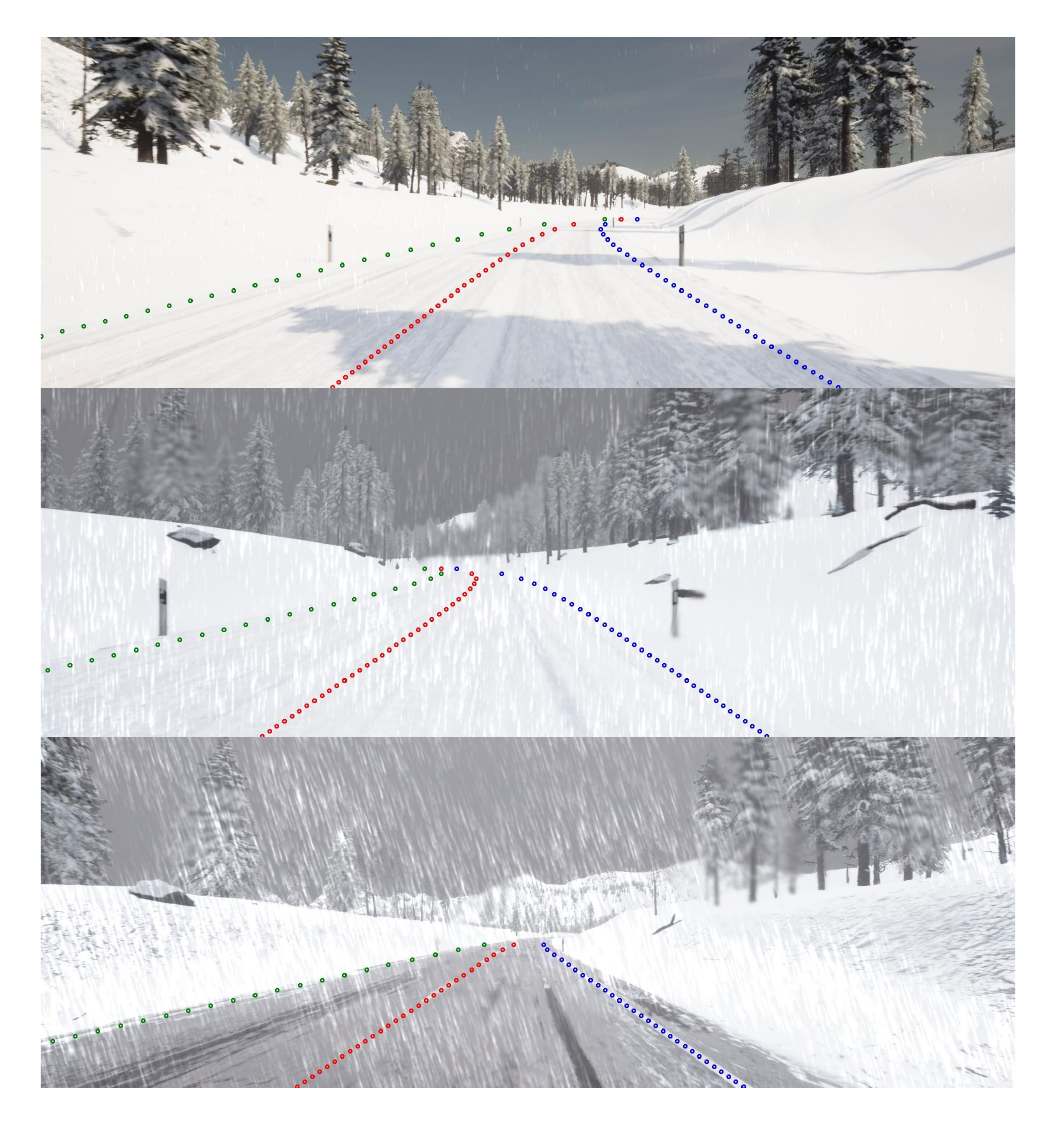}
    \caption{Example images including the snowfall augmentation with varying intensities}
    
    \label{fig:snowfall}
\end{figure}

Lane detection is affected not only by snow covering the road, but also by snowfall. To address this issue, we have incorporated a snowfall augmentation, as discussed in Sec.~\ref{sec:results}. Examples of scenes including the augmented snowfall are shown in Fig.~\ref{fig:snowfall}. The lane detector and delineator-based approach results under snowfall are shown in Tab.~\ref{tab:results_snow_aug}. For the baseline, it can be observed that the results are similar to those without snowfall augmentation. While some methods, such as SCNN, achieved a 3D accuracy of 0.49 with snow, which is an improvement on the results without snowfall, UFLD showed a slight decrease in all metrics. The falling snow led to a slight increase in LSM scores of about 0.02–0.05 for all delineator-based approaches.

Taking into account the additional results, which include snowfall augmentation in the training process, all lane detectors achieve an LSM score greater than 0.90, indicating a completely safe perception. However, a slight decrease by up to 0.10 for 2D accuracy (UFLD) and 0.15 for 3D accuracy (RESA) could be observed.
For delineator-based approaches using Faster R-CNN and YOLO, a significant drop of up to 0.60 in 3D accuracy is evident. Nevertheless, LSM scores of around 0.90 still indicate safe lane detection. For SMOKE, a minor decrease in 3D accuracy and a significant decrease in LSM of up to 0.10 can be observed. Nevertheless, LSM scores ranging from 0.70 to 0.92 indicate that detection can still be considered safe. LiDAR-based detection is almost unaffected by snowfall, showing only minor deviations of 0.02–0.05 in both 3D accuracy and LSM.
In total it can be observed that the snowfall does not show a significant affect on the detection performance of the lane detectors as well as the delineator-based approaches.

\subsection{Realtime Capability}
In order to be suitable for realworld autonomous driving, newly developed methods must satisfy realtime requirements. Hence, we evaluated the computational cost of the delineator-based approach as shown in Sec.~\ref{subsec:delineator-based} using the different detectors.
To allow for a meaningful statement, we report mean processing time $\mu$, standard deviation $\sigma$ and the \SI{95}{\percent} percentile $Q_{95}$ .
\begin{table}[b]
\caption{Results on runtime analysis for delineator-based approach. Results in~\si{\milli\second}.}
    \centering
    \begin{tabular}{lccc }
    \toprule
     & $\mu$ [\si{\milli\second}] & $\sigma$ [\si{\milli\second}] & $Q_{95}$ [\si{\milli\second}] \\
    \midrule
    Faster R-CNN & 2.74 & 0.71 & 4.01\\
    YOLOv5 & 1.89 & 0.41 & 2.49\\
    SMOKE & 2.39 & 0.35 & 2.86\\
    Voxel R-CNN & 0.96 & 0.23 & 1.38\\
    PV-RCNN & 1.16 & 0.23 & 1.55\\
    \bottomrule
    \end{tabular}
    \label{tab:runtime}
\end{table}

For camera-based methods, Faster R-CNN achieves a mean computation time of \SI{2.74}{\milli\second} with a standard deviation of \SI{0.71}{\milli\second} and a \SI{95}{\percent} percentile of \SI{4.01}{\milli\second}. Using YOLOv5 the required time consumption is slightly lower with a mean of \SI{1.89}{\milli\second}, a standard deviation of \SI{0.41}{\milli\second} and a maximum of \SI{2.49}{\milli\second}. SMOKE achieves results between Faster R-CNN and YOLO with a mean of \SI{2.39}{\milli\second}, a standard deviation of \SI{0.35}{\milli\second} and a maximum of \SI{2.86}{\milli\second}.

The LiDAR-based methods only required about \SIrange{30}{50}{\percent} computation time compared to the camera-based methods. With a mean of \SI{0.96}{\milli\second} and \SI{1.16}{\milli\second} for Voxel R-CNN and PV-RCNN, respectively. Considering the standard deviation of \SI{0.23}{\milli\second} for both detectors, it can be stated that the required computation time is nearly constant and with maximum computation times of \SI{1.38}{\milli\second} and \SI{1.55}{\milli\second} in worst case these methods are faster than the camera-based methods in their mean.

The slightly higher computation time of the camera-based methods can be traced back to the required transformation from 2D into 3D coordinates; however, it can be observed that for all detection methods the time for calculating the Bézier curves achieves realtime.

\subsection{Discussion}
Since the effect of delineators in the direct approach with SCNN, RESA and UFLD is negligible, within this section only the results with delineators are compared to the baseline.
Comparing the results with delineators to the baseline it can be observed that the robustness of the lane detection can significantly be improved by a training with adverse weather data. For high and all snow an increase in 2D accuracy between 0.11 (UFLD) and 0.83 for SCNN can be achieved. For low snow the increase is only about 0.40 in 2D accuracy. This can be traced back as in the baseline the training is performed with clearly visible lines which is similar to the low snow environment; hence, the environment to perceive is more similar which already leads to appropriate baseline results. However, the 3D accuracy can be increase by up to 0.28 in high snow for SCNN. Also for the LSM score of SCNN and RESA an increase of about 1.00 can be observed, which corresponds to a completely safe perception of the lanes. Considering snowfall, there was a decrese in accuracy of about 0.10; however, the LSM scores still showed a nearly perfect safety.

For the delineator-based approach using cameras, a significant improvement compared to the baseline can be observed when incorporating winter data into the training process. In the baseline Faster R-CNN and YOLOv5 were not capable to detect any delineators in the baseline. Using SMOKE, LSM scores of up to 0.70 could be achieved already for the baseline, showing a highly safe detection. For the LiDAR-based detection, the results between baseline and training with winter data differs only by about 0.01 in terms of 3D accuracy and LSM. The additional evaluation using snowfall augmentation showed, that snowfall does not affect the perception drastically. Only a minor decrease of some percentage points occurred in 3D accuracy. Compared to the direct approach a significant increase of up to 0.60 in 3D accuracy can be observed for medium snow with PV-RCNN compared to UFLD. For all snow levels and all delineator-based methods an increase of 3D accuracy by over \SI{100}{\percent} can be achieved. Compared to the camera-based methods, the LSM scores are slightly lower but still maintain at a level which corresponds to a completely safe situation. Since the delineators have a distance of about \SI{30}{\metre}, often only one further delineator can be detected while the lane detectors have a higher viewing distance which allows for a higher LSM score. However, for snow-covered roads the vehicle's velocity should be reduced in order to drive safety. This reduction in velocity would then also increase the LSM score as a lower perception distance is required. 

As a summary it can be observed that both the direct and delineator-based approach allow for a realtime applicability, a higher robustness and a better perception of the lanes compared to the lane detectors trained without winter weather data. In terms of safety, the direct approach performs slightly better on the SnowyLane dataset, due to a enhanced perception range. However, the delineator-based approach shows a significantly higher 3D accuracy. In addition, the approach is promising to be robust against to changes in the number of lanes as the number of lanes can be easily determined using Eq.~\eqref{eq:num_lanes} which then allows for a correct shifting of the calculated Bézier curve. For merging lanes this approach can also be applied. Here, only a small deviation at the end of the merging lane will be present.

%%%%%%%%%%%%%%%%%%%%%%%%%%%%%%%%%%%%%%%%%%%%%%%%%%%%%%%%%%%%%%%%%%%%%%%%%%%%%%%%
\section{CONCLUSION \& OUTLOOK}
\label{sec:conclusion}
In this paper we presented SnowyLane, the first large-scale lane detection dataset with snow-covered rural roads and a robust method for lane detection for snow-covered roads in rural environments using parameterized Bézier curves based on infrastructure elements. 
We constructed an extensive benchmark for the SnowyLane dataset with different state-of-the-art lane detectors and showcased the improvement in robustness by incorporating adverse weather data into the training process of the lane detection models. As a baseline, we use data in summer weather conditions to demonstrate the robustness improvement. Using the direct approach, we could increase the result by about 80 percentage points (p.p.) in terms of 2D accuracy, about 10 p.p. for 3D accuracy and up to 1.00 for lane safety metric. In order to achieve a more robust and precise lane detection in snow-covered rural environments we present a novel approach with camera and LiDAR-based object detection for delineators which are standardized in position and dimension. This allows to efficiently derive parameterized Bézier curves for each lane marking. In comparison to the direct approach, this delineator-based approach further improves the perceptual performance by up to 60 p.p. in 3D accuracy.
Due to the easily derivable number of lanes based on knowledge about road construction regulations, the approach is also robust against changes in the number of lanes. Moreover, we demonstrated the realtime capability with a mean calculation time of about \SI{2}{\milli\second} for camera-based methods and about \SI{1}{\milli\second} for LiDAR-based delineator detection.
For further research, the approach will be tested under further adverse weather conditions like rain and fog. Moreover, the benchmark will be extended by further metrics such as F1-score to allow for a broader comparison.

%\newpage

%%%%%%%%%%%%%%%%%%%%%%%%%%%%%%%%%%%%%%%%%%%%%%%%%%%%%%%%%%%%%%%%%%%%%%%%%%%%%%%%

%\addtolength{\textheight}{-2.5cm}

%%%%%%%%%%%%%%%%%%%%%%%%%%%%%%%%%%%%%%%%%%%%%%%%%%%%%%%%%%%%%%%%%%%%%%%%%%%%%%%%

%%%%%%%%%%%%%%%%%%%%%%%%%%%%%%%%%%%%%%%%%%%%%%%%%%%%%%%%%%%%%%%%%%%%%%%%%%%%%%%%
%\section*{APPENDIX}


\section*{ACKNOWLEDGMENT}

This work was supported in part by the German Federal Ministry of Research, Technology and Space (BMFTR) within the
project MANNHEIM-FlexKI under Reference Number 01IS22086H.

%%%%%%%%%%%%%%%%%%%%%%%%%%%%%%%%%%%%%%%%%%%%%%%%%%%%%%%%%%%%%%%%%%%%%%%%%%%%%%%%

\bibliographystyle{IEEEtran} % use IEEEtran.bst style
\bibliography{literature.bib}

\end{document}